\documentclass[runningheads]{llncs}
\usepackage{wrapfig}
\usepackage{graphicx}
\usepackage[margin=1in]{geometry}
\usepackage{multirow}
\usepackage{graphicx}
\usepackage{balance}
\usepackage{xcolor}
\usepackage{colortbl}
\usepackage{amsmath}
\usepackage{comment}
\definecolor{Gray}{gray}{0.65}
\usepackage[ruled,vlined]{algorithm2e}
\usepackage{subcaption}
\begin{document}
\title{HeTriNet: Heterogeneous Graph Triplet Attention Network for  Drug-Target-Disease Interaction}
%
%
\author{Farhan Tanvir\inst{1}\orcidID{0009-0008-8526-8159} \and
Khaled Mohammed Saifuddin\inst{1}\orcidID{0000-0002-0903-937X} \and
Tanvir Hossain\inst{1}\orcidID{0009-0009-2305-6828}  \and
Arunkumar Bagavathi\inst{2}\orcidID{0000-0002-7135-4602}\and
Esra Akbas\inst{1}\orcidID{0000-0002-8817-2442}}
\authorrunning{Tanvir et al.}
%
\institute{Georgia State University, Atlanta GA 30302, USA 
\email{\{ftanvir@,ksaifuddin1@student.,thossain5@student.,eakbas1@\}gsu.edu}
\\
\and
Oklahoma State University,
Stillwater, OK 74075, USA\\
\email{abagava@okstate.edu}}
\maketitle              
\begin{abstract}

Modeling the interactions between drugs, targets, and diseases is paramount in drug discovery and has significant implications for precision medicine and personalized treatments. 
Current approaches frequently consider drug-target or drug-disease interactions individually, ignoring the interdependencies among all three entities. 
Within human metabolic systems, drugs interact with protein targets in cells, influencing target activities and subsequently impacting biological pathways to promote healthy functions and treat diseases. Moving beyond binary relationships and exploring tighter triple relationships is essential to understanding drugs' mechanism of action (MoAs). Moreover, identifying the heterogeneity of drugs, targets, and diseases, along with their distinct characteristics, is critical to model these complex interactions appropriately.
To address these challenges, we effectively model the interconnectedness of all entities in a heterogeneous graph and develop a novel Heterogeneous Graph Triplet Attention Network (\texttt{HeTriNet}). \texttt{HeTriNet} introduces a novel triplet attention mechanism within this heterogeneous graph structure. Beyond pairwise attention as the importance of an entity for the other one, we define triplet attention to model the importance of pairs for entities in the drug-target-disease triplet prediction problem. Experimental results on real-world datasets show that \texttt{HeTriNet} outperforms several baselines, demonstrating its remarkable proficiency in uncovering novel drug-target-disease relationships. 

\end{abstract}
\newpage
\section{Introduction}
\vspace{-2mm}
Understanding the mechanism of actions (MoAs) of drugs is a critical step in drug discovery~\cite{Hauser2017TrendsIG}. However, the traditional approach of high-throughput screening for identifying new drugs is both time-consuming and expensive, taking more than a decade and costing billions of dollars \cite{Hauser2017TrendsIG}. To accelerate this process, computational methods have emerged as valuable tools for leveraging large-scale chemical and genomic data \cite{Chen2018DrugComSD}. Furthermore, in addition to new drug discovery, drug repurposing, or drug repositioning for a disease, is crucial to accelerate this process. It also offers promising avenues for personalized medicine and targeted therapies. 

Recent advancements in machine learning have facilitated the exploration of drugs' MoAs through various learning tasks, such as drug behavior analysis, target activity evaluation, and disease modeling \cite{Eguale2016AssociationOO,Sarker2015PortableAT,ZitnikAL18}. Among these tasks, drug-disease and drug-target prediction have gained significant attention \cite{Ezzat2017DrugTargetIP,Yan2019PredictionOD}. While existing methods have made progress in predicting drug-disease and drug-target relationships, they often treat these tasks as separate entities, overlooking the interconnected nature of drugs, targets, and diseases. The therapeutic effect of drugs on diseases involves their interactions with biological targets, which play a crucial role in disease pathways and overall metabolic system function \cite{Hauser2017TrendsIG}. Therefore, a more comprehensive triple relationship involving drugs, targets, and diseases needs to be considered to capture the interplay between these entities.

Tensor factorization has emerged as a popular approach for inferring missing entries in drug-target-disease tensors and extracting latent structures from high-dimensional data \cite{Chen2019ModelingRD,Wang2015RubikKG}. However, traditional tensor models like CP and Tucker suffer from limitations, including linearity and data sparsity issues. Nonlinear tensor factorization methods have shown promise in capturing the complexities of the data, but they often rely on prior Gaussian processes that are challenging to estimate \cite{Zhe2016DistributedFN}. Moreover, incorporating auxiliary information into tensor models requires tedious feature engineering, making it difficult to handle large-scale healthcare data \cite{Shan2016DeepCW}. NeurTN \cite{Chen2020LearningDD} combines tensor algebra and deep neural networks to learn the intrinsic relationships among drugs, targets, and diseases. By replacing multilinear multiplication with a neural network, NeurTN can capture nonlinear dependencies among tensor data. While NeurTN offers promising capabilities for understanding drugs' MoA, it is important to consider some potential drawbacks, including interpretability challenges of black-box models, the need to assess generalization to new data, and the importance of comprehensive comparisons with existing methods.

The activity of drug-target-disease association involves drugs and various biomedical entities. Different types of drugs interact with different entities, resulting in different events. A question might arise- how do we model these complete, enriched data? Networks, often representing real-life systems, are graphs that capture the complex structure of interactions between related objects. Networks exist in multiple disciplines, such as social networks \cite{akbas2017truss,akbas2020proximity}, citation networks \cite{tanner2019paper,lopes2010collaboration}, and biological networks~\cite{Tanvir2021PDDI,Bumgardner2021DrugDrugIP,Tanvir2023PredictingDI,Saifuddin2022HyGNNDI,Saifuddin2023SeqHyGANSC}. To represent various entities and their disparate interactions, heterogeneous Information Network (HIN)\cite{Shi2017ASO} is defined. Heterogeneous graphs provide a powerful framework for representing the diverse entities and interactions involved in drug discovery. In these graphs, nodes represent different types of entities, such as drugs, proteins, pathways, chemical substructures, ATC codes, and diseases, while edges capture the interactions between these entities.

In this paper, we consider the drug repurposing problem as a triplet prediction task while capturing the intricate relationships among drugs, proteins, and diseases. For this problem, we propose a novel \texttt{He}terogeneous Graph \texttt{Tri}plet Attention \texttt{Net}work (\texttt{HeTriNet}), which leverages the power of heterogeneous graphs. By leveraging the rich information encoded in the heterogeneous graph, \texttt{HeTriNet} addresses the limitations of traditional tensor models and offers a more effective approach for modeling and analyzing complex systems. \texttt{HeTriNet} employs a triplet attention mechanism to capture higher-order interactions among drug-target-disease triplets in the heterogeneous graph. This attention mechanism enables the model to learn complex patterns and relationships within the data, leading to improved predictive performance and a deeper understanding of biological and chemical processes. Our main contributions are as follows:

\begin{itemize}
\item \textbf{Addressing drug-target-disease associations using a heterogeneous graph neural network:} We propose a novel approach that models the complex interactions between drugs, targets, and diseases using a heterogeneous graph neural network (HGNN). By incorporating different types of nodes and edges, our approach effectively captures the rich information embedded in the interactions between these entities, leading to improved prediction performance.

\item \textbf{Introducing \texttt{HeTriNet} model:} We develop a novel model, \texttt{HeTriNet}, that extends traditional graph neural networks by incorporating a novel triplet attention mechanisms. No prior work in GNN and HGNN has explored triplet-wise attention mechanisms. This allows the model to capture higher-order triplet-wise interactions and relationships in the heterogeneous graph. This attention mechanism enables the model to focus on the most relevant connections and interactions, improving its generalization and predictive accuracy.
\item \textbf{Extensive Experiments:} We conduct extensive experiments to show the effectiveness of our model on two different datasets. We also
compare the proposed \texttt{HeTriNet} with the state-of-the-art baseline models. The results with different accuracy measures show that our method significantly surpasses the baseline models. In addition, different case studies denote that our model's predictions can be validated by different datasets as well as external literature evidence.

\end{itemize}

The structure of this paper is outlined as follows. In this section, we explain drug repurposing and drug-target-disease interaction. In addition, we explore related works in Section \ref{works}. Section \label{method} describes how we create a heterogeneous graph. Moreover, we elaborate on the \texttt{HeTriNet} model, consisting of an encoder-decoder framework. Furthermore, we describe our experiments and illustrate our results in Section \label{experiment}. Finally, we conclude in Section \label{conclusion}.

\vspace{-4mm}
\section{Related Works}\vspace{-3mm}
\label{works}
In this section, we present an overview of relevant research in the field of computational predictions of drugs, targets, and diseases, with a specific focus on triplet prediction for high-dimensional structured data.
\\\textbf{Modeling drug-target-disease}
Treating human diseases caused by complex biological processes involves the activities of many biological entities, such as drugs and targets. Computational pharmacology methods aim to find associations among these entities and understand drug mechanisms of action (MoAs) \cite{Hauser2017TrendsIG}. Numerous approaches have been developed to predict drug-disease and drug-target relationships using computational methods. Network-based inference models, such as bipartite networks of drugs and diseases (or targets), have been widely explored \cite{Chen2012DrugtargetIP,Ezzat2017DrugTargetIP,Yamanishi2008PredictionOD}. These models utilize machine learning algorithms, including random walk, matrix factorization, and support vector machine, to predict novel interactions between drugs and diseases (or targets). Incorporating existing medical knowledge, such as omics data, from web sources has also been investigated to gain a better understanding of complex human metabolic systems\cite{Chen2017AFA,Nascimento2016AMK,Zheng2013CollaborativeMF,ZitnikAL18}. Multiple kernel learning methods have been developed to predict drug-target relationships by integrating various similarities of drugs and targets \cite{Nascimento2016AMK}. Other models, such as Decagon, model drug polypharmacy side effects by incorporating additional drug-drug and protein-protein interactions \cite{ZitnikAL18}.

However, one significant limitation of most existing approaches is that they treat drug-disease and drug-target predictions as separate tasks, overlooking the relationships among drugs, targets, and diseases. Instead of modeling the triple relationship $<$drug, target, disease$>$, they focus on binary relationships or pairwise interactions. This limitation restricts the comprehensive understanding of drug mechanisms and their effects on diseases.
\\\textbf{Triplet Prediction}
Triplet prediction has been applied in various domains, including drug repurposing and natural language processing. Zhang et al. \cite{Zhou2021TripletAR} introduced a novel attention mechanism based on transformers that capture relationships between three entities (query, key, and value) and improve language understanding and generation. In natural language processing, triplet prediction has been applied to tasks like relationship extraction, where models utilize sentence-level attention and entity descriptions to predict relationships between entity triplets in text \cite{Ji2017DistantSF}. In computer vision, triplet prediction techniques have been utilized for tasks such as face recognition and person re-identification. Schroff et al. \cite{Schroff2015FaceNetAU} proposed a deep convolutional network that learns embeddings by optimizing the similarity between images in the same class while maximizing the distance between images from different classes. In-person re-identification, Hermans et al. \cite{Hermans2017InDO} introduced an improved triplet loss function for deep learning models, enabling them to learn discriminative embeddings.

In the drug repositioning domain, Zitnik and Zupan \cite{Zitnik2013DataFB} presented a data fusion method based on collective matrix factorization to model drug-target-disease associations. Their method integrates multiple sources of information by jointly factorizing multiple matrices, capturing the interactions between drugs, targets, and diseases. In the evolving landscape of drug discovery, a recent research endeavor by Qu et al. \cite{EGNN} explores the interdependence of drugs, targets, and diseases through event-graph modeling. This innovative approach employs the concept of an event to describe the interdependence of drug-target-disease as a complete semantic unit. Zhang et al. \cite{Chen2020LearningDD} developed a neural tensor network model that integrates drug, target, and disease information to predict the drug-target-disease triplet relationship. This approach allows for the simultaneous identification of potential drug targets and the diseases they may treat. In a similar vein, Chen and Li \cite{Chen2019ModelingRD} focused on predicting drug-target-disease relationships using tensor factorization techniques, which are capable of capturing the latent structure in high-dimensional data. By incorporating heterogeneous information from various web sources, their approach provided a more comprehensive understanding of the underlying biological processes involved in drug-target-disease interactions.

In the context of drug-target-disease association prediction, the application of heterogeneous graph neural networks (HGNNs) remains largely unexplored. Additionally, the incorporation of triplet attention mechanisms within both GNNs and HGNNs for capturing the relationships between triplets has not been previously investigated. To bridge this gap, we propose a novel framework, the Heterogeneous Graph Triplet Attention Network (\texttt{HeTriNet}). By leveraging the power of HGNNs and incorporating triplet attention mechanisms, our method aims to determine the complex relationships and dependencies among drugs, targets, and diseases in a unified framework. This innovative approach provides a unique perspective and contributes to advancing the field of drug discovery.
\vspace{-4mm}
\section {Methodology}\vspace{-3mm}
\label{method}
Given a triplet includes a drug, a target, and a disease, our goal is to predict whether the triplet has an interaction or not. In this paper, we propose a novel approach that leverages the power of heterogeneous information networks (HIN) and introduces the concept of triplet attention. To achieve this, we develop the Heterogeneous Graph Triplet Attention Network (\texttt{HeTriNet}), which employs an end-to-end encoder-decoder architecture. The encoder integrates a triplet attention mechanism to determine the significance of pairs (e.g., target-disease) for the other entity (e.g., drug) while learning embeddings of all entities and also triplets. Moreover, \texttt{HeTriNet} incorporates a decoder that learns and predicts the interaction between entities of triplets. System Architecture of \texttt{HeTriNet} is outlined in Figure~\ref{fig:sys_arch}. Our proposed model consists of the following steps:
\begin{figure*}[t]
    \centering
    \includegraphics[width=\textwidth]{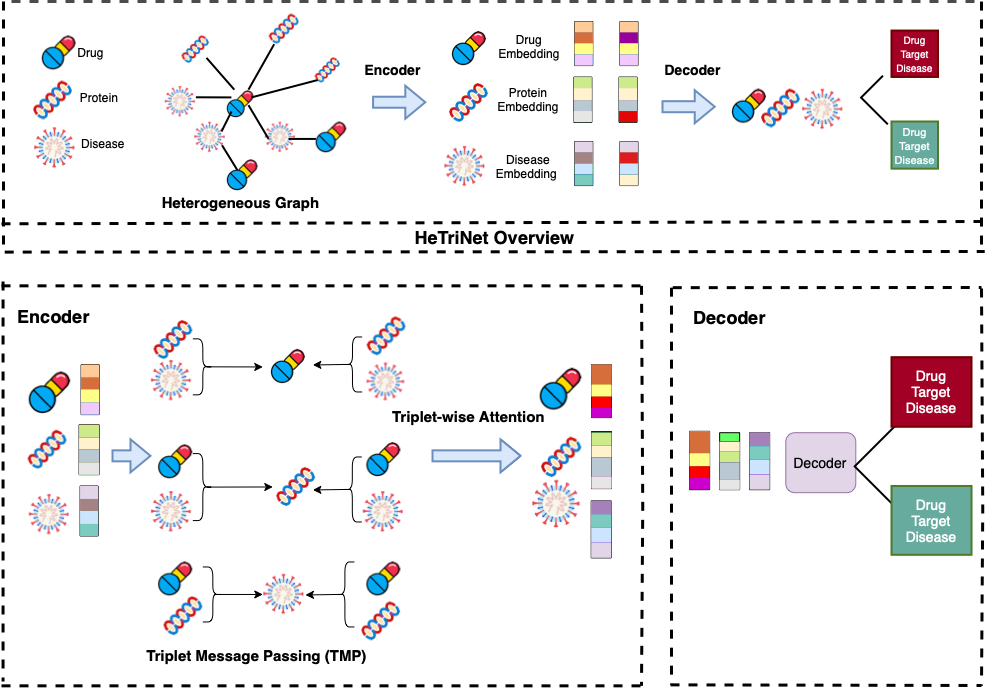}
   \caption{System Architecture of \texttt{HeTriNet}}
    \label{fig:sys_arch} \vspace{-6mm}
\end{figure*}
\vspace{-2mm}
\begin{enumerate}
\item Heterogeneous Graph Construction \& Node's Feature Extraction
\item Heterogeneous Graph Triplet Attention Network Architecture
\begin{itemize}
\item Encoder: Triplet Attention-based Node Representation learning
\item Decoder: Drug-target-disease Interaction Prediction
\end{itemize}
\end{enumerate}
\vspace{-4mm}
\subsection{Heterogeneous Graph Construction \& Node's Feature Extraction}\vspace{-2mm}
The first step in our approach is to construct a heterogeneous graph that captures the complex relationships among drugs, proteins (targets), and diseases. The graph consists of three types of nodes: drug, protein, and disease. We establish edges between these nodes based on known drug-target interactions and drug-disease associations. This construction allows us to represent the rich interactions and dependencies between different entities in the graph.

One of the biggest strengths of graph neural networks (GNNs) is including node features in the learning process. In a general GNN architecture, initialized node features are used to generate enriched and effective node embedding through a message-passing mechanism. Therefore, we need features of nodes to give to the GNN model in the next step. In the feature extraction step, we extract features for each node in the graph. Specifically, we focus on the chemical substructures of drugs and targets, represented as SMILES strings and Amino Acid sequences, respectively. Disease node features are represented using one-hot encoding. To create drug and target features, we employ the ESPF~\cite{espf} algorithm. ESPF decomposes the SMILES string and Amino Acid sequence into frequent substructures, selecting the most significant ones based on a frequency threshold. These substructures provide informative features for the drugs and targets, which is utilized in the subsequent steps of the \texttt{HeTriNet} model.

\vspace{-4mm}
\subsection{Heterogeneous Graph Triplet Attention Network Architecture}
\vspace{-2mm}
The core goal of our research is to address the challenge of predicting drug-target-disease interactions accurately. To achieve this, our model \texttt{HeTriNet} leverages the rich information in heterogeneous networks and captures the complex relationships among drugs, targets, and diseases. The model is trained using an end-to-end approach to predict drug-target-disease interactions. The model is responsible for aggregating information from neighboring nodes and learning higher-order relationships in the graph using a triplet attention mechanism, the critical component of our model. The triplet attention mechanism calculates attention coefficients based on the features of all three nodes in a triplet ($i$, $j$, and $k$), where $i$ is the central node, and $j$ and $k$ are neighboring nodes. These coefficients are used to weigh the importance of neighboring node pairs for the central nodes in each triplet when aggregating information. The attention mechanism is applied to all generated triplets in the graph, enabling the model to capture complex, interconnected relationships among the different node types. Afterward, the encoded representations of nodes are obtained by aggregating information from neighboring nodes, weighted by the attention coefficients. This process is repeated for each layer of the \texttt{HeTriNet} model, allowing the model to learn increasingly complex patterns and dependencies across multiple layers.

 Our proposed model consists of the following steps:

\begin{itemize}
    \item Encoder: Triplet Attention-based Node Representation learning
    \item Decoder: Predicting Drug-Target-Disease Interactions
    
\end{itemize}

\textbf{\texttt{i. Node Representation learning (Encoder):}}
The encoder component of our model focuses on learning informative node representations. In a heterogeneous graph, nodes and edges belong to different types, and each type of node has its own distinct feature space. To effectively learn informative node representations, we must align these diverse feature spaces into a common one. This enables meaningful comparisons and interactions among nodes of different types. To overcome this challenge, we introduce a type-specific transformation matrix $M$, which projects the features of different nodes into a common feature space as follows: $ h_i^{'}  = M \circ h_i $

In a regular Graph Convolutional Network (GCN), message passing between nodes occurs in pairs, where neighboring nodes pass messages to each other. However, real-world data often involves more intricate dependencies between nodes that extend beyond pairwise relationships. Pairwise message passing, where one node passes the message to another, limits the capturing of this complex dependency. This limitation is significantly pronounced in our scenario, where mere message passing between drugs and targets, or drugs and diseases, may not capture the complex relations inherent in drug-target-disease triplets. As a solution to this limitation, we define \textbf{Triplet Message Passing} function (TMP) to learn the node representation capturing the complex relations between nodes. For a central node of type $i$, we define its neighbors as node pairs ($N_i$), comprising node pairs of type $j$ and $k$. We pass messages from these neighbor pairs to the central node. The goal is to capture complex relationships, intricate patterns, and essential interactions within drug-target-disease data, which can be expressed as:

\begin{equation}
\label{eq1}
\begin{split}z_i^l= TMP(z_i^{l-1}, N_i)\end{split}
\end{equation}

For one central node, there are several node pairs as the neighbors. However, it is essential to note that not all neighbor pairs are equally crucial for the central node. Message passing should consider these varying levels of importance.  To assess the importance of neighbor pairs for a central node, we design a novel \textbf{Triplet-wise Attention} mechanism.  Given a central node $i$ and its neighboring nodes $(j, k)$, we employ triplet-wise attention on these nodes to determine their relative importance for the target node. The attention mechanism calculates attention coefficients $e_{ijk}$ based on the features of all three nodes in a triplet and is represented as follows:

\begin{equation}
\label{eq2}
\begin{split}
e_{ijk} & = a (h_i^{'}, h_j^{';}, h_k^{';} ) \\
& = LeakyRELU( NN ( h_i^{'} || h_j^{';} || h_k^{';} ) ) \\
\end{split}
\end{equation}

In Eq.~\ref{eq2}, $a$ denotes triplet-wise attention mechanism and $||$ denotes the concatenate operation. In the attention mechanism, we employ a neural network, denoted as $NN$. This neural network is designed to capture complex relationships and dependencies among the features of the nodes in a triplet. Additionally, to capture the nonlinear dependencies among drug-target-disease data, we apply the LeakyReLU activation function. LeakyReLU is chosen for its ability to introduce nonlinearity in the model, allowing it to capture complex relationships critical for accurately predicting drug-target-disease interactions.

It is vital to make attention coefficients easily comparable across different nodes. Therefore, the attention coefficients are then normalized using a Softmax function. This step ensures that the model appropriately weighs the attention of each neighbor when aggregating information. The normalized attention coefficients $\alpha_{ijk}$ determine the distribution of attention among various neighbor pairs when aggregating information from neighboring nodes. 

\begin{equation} \label{eq3}
\begin{split}
\alpha_{ijk} & = softmax_j(e_{ijk}) \\
& = \frac{\exp(e_{ijk})}{\sum_{l,m \in N(i)} \exp(e_{ilm})  } \\
\end{split}
\end{equation}

During the triplet message passing process, it is imperative to consider the message from neighbor pairs. To generate messages from pairs, we concatenate the representations of the nodes within the pair. We then pass this concatenated feature vector of size $2d$ through a single-layer feedforward neural network to transform it into a feature vector of size $d$. After multiplying each pair message with calculated pair attention, we sum all of them to create the final message from neighbor pairs. Then, the obtained neighbor message is combined with the central node representation with a self-attention model. So, the \textbf{TMP} in Eq~\ref{eq1} is defined as follows:

\begin{equation} \label{eq4}
\begin{split}
z_{i} & = \delta( h_i^{'} + W * \sum_{j,k \in N(i)} (\alpha_{ijk} \circ NN ( h_j^{'} || h_k^{'} ) ) ) \\
\end{split}
\end{equation}

In order to incorporate self-attention, we use a trainable parameter $W$, which determines the weight or importance of node $i$'s embedding in the aggregation process. The purpose of $W$ is to control the combination of node $i$'s features with the aggregated features of its neighboring nodes. By adjusting the value of $W$, the model captures self-attention and measures the influence of node $i$ on the overall representation.

To capture more complex patterns and relationships, we employ multi-head attention. This mechanism introduces parallel attention mechanisms that can focus on different aspects of the data, enhancing the model's ability to learn intricate patterns. In multi-head attention, multiple attention mechanisms (K) are used individually to transform the features, and the outputs are concatenated (symbolized as $||$) to obtain the final representation.

\begin{equation} \label{eq5}
\begin{split}
z_{i} & = ||_{k=1}^K \delta( h_i^{'} + W * \sum_{j,k \in N(i)} (\alpha_{ijk}  \circ NN ( h_j^{'} || h_k^{'} ) ) ) \\
\end{split}
\end{equation}

Through the integration of these equations and steps, our model effectively acquires informative node representations within a heterogeneous graph. This ensures that it captures complex relationships, intricate patterns, and crucial interactions vital for predicting drug-target-disease interactions.
\\{\texttt{ii. Predicting Drug-Target-Disease Interactions (Decoder):}} 
The decoder component of our model is responsible for predicting the likelihood of interactions between drugs, targets, and diseases based on the encoded representations obtained from the encoder. Decoder, in particular, assigns a score to drug, target, and disease triplet ($v_i$,$v_j$,$v_k$). It expresses how likely it is that drug $v_i$. target $v_j$, and disease $v_k$ are interacting. The corresponding entities' features are concatenated and passed through a multilayer perceptron (MLP).
\begin{equation}
\label{eq6}
\begin{split}
pred_{x,y,z} = MLP ( z_x || z_y || z_z )
\end{split}
\end{equation}

The multilayer perceptron (MLP) generates a prediction score, $Y'$, ranging from 0 to 1. A score close to 1 indicates a high likelihood of interaction among the triplets, whereas a score close to 0 indicates that interaction is less likely.
\vspace{1mm}
\\{\texttt{iii. Model Optimization:}}
We adopt pairwise learning methods to optimize the model parameters. The objective is to ensure that observed triplets are predicted with higher scores than unobserved ones. This is achieved by minimizing the discrepancy between the predicted and ground truth labels using a loss function cross-entropy. 
 
\begin{equation}
\label{eq7}
\begin{split}
L= \sum_{i,j,k \in \mathcal(D^{+})} \sum_{i^{'},j^{'},k^{'} \in \mathcal(D^{-})} max(0,1 + f(i^{'},j^{'},k^{'}) - f(i,j,k) )
\end{split}
\end{equation}

Here, $f{}$ denote the predictive function in Eq ~\ref{eq7}. $\mathcal(D^{+})$ denotes the set of positive triplets and $\mathcal(D^{-})$ denotes the set of negative triplets corresponding to $\mathcal(D^{+})$ by sampling from unobserved elements. For each positive training triplet $(i,j,k)$,  we randomly sample a negative training triplet $(i^{'},j^{'},k^{'})$ and optimize the model parameters based on the pairwise learning principle.

By applying the \texttt{HeTriNet} model and optimizing its parameters, we aim to predict drug-target-disease interactions accurately. The model's encoder-decoder architecture, incorporating triplet attention mechanisms, enables it to capture intricate relationships in heterogeneous networks. Our model's ability to learn informative node representations and make precise predictions has the potential to advance drug discovery and aid in understanding the mechanisms of action.
\vspace{-4mm}
\section{Experiment}\vspace{-3mm}
\label{experiment}

To evaluate our \texttt{HeTriNet} model, we conduct experiments involving negative sampling and random dataset splitting into train and test sets. Our performance assessment included accuracy, Precision, F1-score, AUROC, and the commonly used top-n metric hit@n. This section summarizes our experimental parameters, evaluation protocols, and analysis of results.
\vspace{-4mm}
\subsection{Datasets, Parameter Settings \& Baselines}
\vspace{-2mm}
Our study relies on data from two essential sources, CTD and DrugBank, offering insights into drug-related information. We conduct experiments using two dataset configurations. One uses data exclusively from DrugBank, encompassing details about drug-target interactions and drug-disease associations. The other configuration integrates information from DrugBank (concerning drug-target interactions) with data from CTD (providing drug-disease associations). This integration provides a comprehensive view of $<$drug, target, disease$>$ triplets. In the remaining sections, DrugBank dataset and the combined Dataset will be referred to as DB and DB\&C, respectively. Table~\ref{tab1} summarizes the key characteristics of the nodes and edges in our heterogeneous graph, forming the basis of our analysis.
\begin{table*}[t]
    \centering
      \caption{Statistics of Dataset}
      \begin{tabular}{|c|c|c|c|c|c|} \hline
\cellcolor{gray!60} \textbf{Dataset} &
\cellcolor{gray!60} \textbf{\# of Drugs} & \cellcolor{gray!60} \textbf{\# of Targets} & \cellcolor{gray!60} \textbf{\# of Diseases}& \cellcolor{gray!60} \textbf{Triplets}\\
\hline
DrugBank (DB)  & 531 & 836 & 279 & 27,238
        \\         \hline
DrugBank and CTD (DB\&C) & 450 & 708 & 1, 267 & 175, 288
\\         \hline
        \end{tabular}
        \label{tab1} \vspace{-2mm}
\end{table*}

The datasets are split into random training (80\%) and testing (20\%) subsets, with five iterations for robust evaluation. Model parameters are optimized using the Adam optimizer with a learning rate of 0.005. To prevent overfitting, dropout and Xavier uniform initialization are applied. Training includes 2000 epochs with early stopping after 200 consecutive epochs without validation loss improvement.

Negative test triplets are generated by negative sampling. To evaluate the performance of our model, we employ the widely-used top-n metric called hit@n and NDCG@n to evaluate recommendation performance, as proposed by \cite{He2017NeuralCF,Bordes2013TranslatingEF}. This metric assesses whether a test triplet is present within the top-n ranked list, while NDCG@n is a position-aware metric that assigns larger weights on higher positions. To generate the top-n ranked list, we use the model's prediction scores for each test triplet $(i, j, k)$. Based on these scores, we rank the triplets in descending order, placing the most likely interactions at the top of the list. The top-n ranked list refers to the subset of triplets that are considered the most probable interactions according to their scores. For instance, we generate 100 random negative samples for each test triplet (i, j, k) and calculated hit@n based on the rankings of these triplets. In addition to hit@n, we also utilized other evaluation measures such as F-1 score, Precision, Recall, ROC-AUC, and AUPR.

We evaluate our model by comparing its performance with different types of state-of-the-art models. We categorize the baseline models into the following groups.

\begin{itemize}
\item \textit{Tensor Decomposition Methods:} CP and Tucker are popular tensor models with diverse variants being successfully applied in health data analysis\cite{Kolda2009TensorDA}. They both adopt multilinear assumptions.

\item \textit{Attention-based Methods:} We use transformer, a powerful deep learning architecture that captures complex relationships and patterns in the data. It utilizes self-attention mechanisms to effectively learn and represent the interactions between drugs, targets, and diseases.

\item \textit{Graph Neural Network (GNN):} We use GNN architectures on our graph to learn the representation of nodes. We select three common GNN-based methods: GIN\cite{Xu2018HowPA}, GAT\cite{Velickovic2018GraphAN}, and GraphSAGE \cite{Hamilton2017InductiveRL}. 

\item \textit{Heterogeneous Graph Neural Network (HGNN):} We use commonly used HGNN model heterogeneous graph transformer (HGT) \cite{Hu2020HeterogeneousGT} on our heterogeneous graph to learning representation of nodes.

\item \textit{NeurTN:} Neural Tensor Network (NeurTN)\cite{Chen2020LearningDD}, which combines tensor algebra and deep neural networks, which offers a more powerful way to capture the nonlinear relationships among drugs, targets, and diseases. Both NeurTN and \texttt{HeTriNet} combine drug-target and drug-disease interactions from DrugBank and CTD.

\end{itemize}
\vspace{-2mm}

\vspace{-4mm}
\subsection{Comparison with baselines}
\vspace{-2mm}
\begin{table*}[t]
\centering{
\caption{ Comparing performance of \texttt{HeTriNet} with other baseline models on DB }
\label{tab3}
\small
\begin{tabular}{|c|c| c c c c c | }
\hline
 \cellcolor{gray!60}\textbf{Model}
 & \cellcolor{gray!60}\textbf{Method}
 & \cellcolor{gray!60}\textbf{F-1 Score}
 & \cellcolor{gray!60}\textbf{Precision}
 & \cellcolor{gray!60}\textbf{Recall}
 & \cellcolor{gray!60}\textbf{ ROC-AUC}
  & \cellcolor{gray!60}\textbf{AUPR}\\ 
\hline

 & TD & 47.00 & 48.51 & 45.59 & 49.19 & 48.98 \\

Tensor-based & CPD & 52.91 & 52.19 & 56.19 & 49.84 & 50.06 \\
\hline
Attention-based & Transformer & 52.31 & 62.96 & 51.18 & 60.62 & 59.46 \\
\hline

& GraphSAGE & 72.24 & 61.3 & 83.94 & 66.4 & 59.92 \\				
GNN-based & GIN & 74.06 & 71.18 & 77.2 & 73.08 & 66.31 \\

& GAT & 72.63 & 62.65 & 82.34 & 67.64 & 60.92 \\
\hline

& HGT & 80.44 & 82.71 & 79.41 & 83.13 & 83.32 \\

HGNN-based & \texttt{HeTriNet} & \textbf{86.31}& \textbf{88.43} & \textbf{84.34}& \textbf{93.46}& \textbf{93.07} \\
\hline
\end{tabular}} \vspace{-4mm}
\end{table*}
\begin{table*}[t]
\centering{
\caption{ Comparing performance of \texttt{HeTriNet} with other baseline models on DB\&C }
\label{tab4}
\small
\begin{tabular}{|c|c| c c c c c | }
\hline
 \cellcolor{gray!60}\textbf{Model}
 & \cellcolor{gray!60}\textbf{Method}
 & \cellcolor{gray!60}\textbf{F-1 Score}
 & \cellcolor{gray!60}\textbf{Precision}
 & \cellcolor{gray!60}\textbf{Recall}
 & \cellcolor{gray!60}\textbf{ ROC-AUC}
  & \cellcolor{gray!60}\textbf{AUPR}\\ 
\hline
 & TD &  53.17 & 62.45 & 47.86 & 60.65 & 62.04 \\

Tensor-based & CPD & 57.23 & 63.72 & 52.19 & 59.82 & 60.76\\
\hline

Attention-based & Transformer & 83.05 & 85.24 & 81.09 & 82.04 & 75.36\\
\hline

& GraphSAGE & 83.31 & 78.97 & 90.04 & 75.98 & 70.72\\

GNN-based & GIN & 83.98 & 78.16 & 83.49 & 76.45 & 71.52\\

& GAT & 85.17 & 82.44 & 82.11 & 83.76 & 77.15\\
\hline

& HGT & 85.22 & 87.08 & 84.55 & 87.98 & 84.07\\

HGNN-based & \texttt{HeTriNet} & \textbf{90.91} & \textbf{93.12} & \textbf{89.88} & \textbf{98.01} & \textbf{97.75}\\
\hline
\end{tabular}} \vspace{-2mm}
\end{table*}
In this study, we conduct a comprehensive performance analysis of \texttt{HeTriNet} in comparison to a selection of state-of-the-art baseline models. To assess the efficacy of these models, we employed a diverse set of performance metrics. Specifically, we report the results for F-1 Score, Precision, Recall, ROC-AUC, and AUPR in Table~\ref{tab3} for DB and Table~\ref{tab4} for DB\&C. In both tables, we refer to the tensor-based baselines, tucker decomposition, and CP decomposition as TD and CPD, respectively. Our model, \texttt{HeTriNet}, outperforms all baseline models for both datasets, showcasing its exceptional predictive capabilities. For instance, on DB, \texttt{HeTriNet} achieved impressive F-1 scores, ROC-AUC, and AUPR of 86.31\%, 93.46\%, and 93.07\%, which represented a substantial improvement over the best-performing baseline, HGT, with F-1 score, ROC-AUC, and AUPR of 80.44\%, 83.13\%, and 83.32\%, respectively.

\begin{figure*}[t]
    \centering    \includegraphics[width=\textwidth]{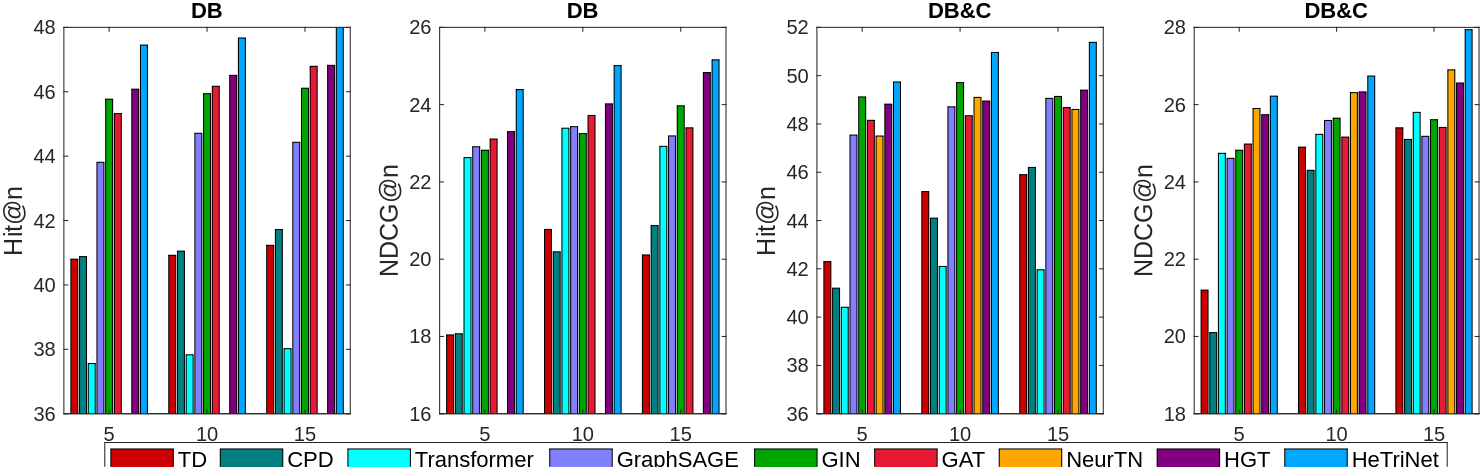}
   \caption{Evaluation of top-n performance for \texttt{HeTriNet} and other baseline models in terms of a) Hit@n and b) NDCG@n on DB and DB\&C}
    \label{fig:Result} \vspace{-6mm}
\end{figure*}

In addition to these performance metrics, we adopt the top-n metrics, Hit@n and NDCG@n, as illustrated in figure~\ref{fig:Result}. In figure~\ref{fig:Result}, These metrics are particularly critical in the context of triplet prediction, as they assess the ranking quality of the model's predictions. Notably, \texttt{HeTriNet} showcased its superiority in terms of ranking quality. It achieved remarkable results in both Hit@n and NDCG@n, significantly outperforming its competitors. For example, it attained a Hit@15 score of 50.11\% and NDCG@15 of 27.36\%, representing substantial enhancements over the top-performing baseline model, HGT.

For a more in-depth analysis, it is worth noting that tensor-based models, like Tucker decomposition and Canonical Polyadic (CP) decomposition, utilize multi-dimensional arrays to capture complex interactions. While these models demonstrate commendable performance in specific metrics, such as F-1 score and Precision, they tend to underperform in terms of Recall and ranking quality. In contrast, attention-based models leverage attention mechanisms to weigh the significance of different input elements, resulting in elevated Precision, Recall, and ranking quality. Furthermore, models like NeurTN combine the strengths of both tensor-based and attention-based models. Although we couldn't obtain accurate results for F-1 score, Precision, and Recall or perform experiments for NeurTN on the DrugBank dataset, its focus on top-n matrices and data combination showcases its unique strengths. The success of GNN and HGNN-based models is also noteworthy, consistently securing F-1 scores surpassing 70\%. Notably, models like HGT stand out as an exemplary baseline, emphasizing the fundamental role of graph structural information in predictive modeling.




The commendable performance of GNN and HGNN-based models underscores the pivotal role played by graph structural information in this context. With their capacity to analyze graph structure data, these models have garnered significant interest due to their effectiveness in graph data analysis. GNNs, in particular, have been instrumental in representing interactions between graph nodes and capturing graph dependence through message passing. Moreover, \texttt{HeTriNet}'s consistent superiority across a range of performance metrics, coupled with its exceptional ranking quality, firmly establishes it as a transformative model in this field. \texttt{HeTriNet}'s distinctive contributions include its holistic approach to drug-target-disease associations, leveraging the power of HGNN, and introducing an innovative triplet-wise attention mechanism.


\vspace{-4mm}
\subsection{Prediction and Validation of Novel Triplets}
\vspace{-2mm}

One of the key objectives of our study is to evaluate the effectiveness of \texttt{HeTriNet} in predicting drug-target-disease interactions using real-world datasets. To assess the model's performance, we employ a rigorous process that involves comparing its predictions with data from two distinct datasets.

\begin{table*}[t]
    \centering
      \caption{Novel Triplet Predictions by \texttt{HeTriNet} FROM DB\&C}
      \begin{tabular}{|c|c|c|c|c|c| }
      \hline
       \cellcolor{gray!60}Drug & \cellcolor{gray!60}Target & \cellcolor{gray!60}Disease & \cellcolor{gray!60}DB\&C Label & \cellcolor{gray!60}Prediction & \cellcolor{gray!60}DB Label \\
       \hline

Carbamazepine & NR1I2-HUMAN & Osteoporosis & 0 & 0.99 & 1 \\

Testosterone & ERR3-RAT & Myocardial infarction & 0 & 0.98 & 1 \\

Nefazodone & DRD2-HUMAN & Schizophrenia & 0 & 0.97 & 1 \\

Raloxifene & ERR3-RAT & Obesity & 0 & 0.93 & 1 \\

Fenofibrate & MMP19-HUMAN & Psoriatic arthritis & 0 & 7e-09 & 0 \\
    
        \hline
        \end{tabular}
        \label{tab7} \vspace{-4mm}
\end{table*}

In our evaluation, we begin by selecting specific triplets from Dataset DB\&C. These particular triplets lack any interaction data within DB\&C, yet they possess relevant association information in DB. To ensure the reliability of our validation, we adopt a comprehensive approach. We train \texttt{HeTriNet} using the DB\&C dataset while ensuring that the selected triplets are exclusively present in the test set of DB\&C. This careful curation of the test set minimizes any potential bias in our evaluation. The next step involves collecting and analyzing the predicted scores generated by our model for these selected triplets. The results, as depicted in Table~\ref{tab7}, reveal an intriguing pattern. For the first four triplets, where the DB\&C labels indicate no interaction, \texttt{HeTriNet} produces predicted scores that consistently exceed the 90\% threshold. This observation strongly suggests that these triplets are highly likely to exhibit interactions despite the absence of explicit interaction data in DB\&C. To further solidify our findings, we cross-check our predicted scores with the information contained in DB. Remarkably, this comparison confirms the existence of interactions between these four drug pairs, reinforcing the predictive power of our model.

\begin{table*}[t]
    \centering
      \caption{Novel Triplet Predictions by \texttt{HeTriNet} FROM DB}
      \begin{tabular}{|c|c|c|c|c|c| }
      \hline
       \cellcolor{gray!60}Drug & \cellcolor{gray!60}Target & \cellcolor{gray!60}Disease & \cellcolor{gray!60}DB Label & \cellcolor{gray!60}Prediction & \cellcolor{gray!60}DB\&C Label \\
       \hline

Cyclobenzaprine & 5HT2C-HUMAN & Muscle Spasm & 0 & 0.99 & 1 \\

Cyclobenzaprine & AA2AR-HUMAN & Gout & 0 & 0.98 & 1 \\

Imipramine & ADA1D-HUMAN & Interstitial Lung Disease & 0 & 0.97 & 1 \\

Quetiapine & HRH1-HUMAN & Schizophrenia & 0 & 9.9e-10 & 0 \\

Verapamil & CAC1S-HUMAN & Cluster headache & 0 & 5e-07 & 0 \\
    
        \hline
        \end{tabular} 
        \label{tab8}\vspace{-4mm}
\end{table*}

We expand our validation process by selecting another set of five drug triplets from DB. These triplets lack interaction information within DB but contain such data in DB\&C, as highlighted in Table~\ref{tab8}. For this set, we tailor our approach by training \texttt{HeTriNet} using DB data, thus specializing the model to this unique dataset. Subsequently, we validate the predicted scores by cross-referencing them with DB\&C. This comprehensive validation process serves to underscore the reliability and generalizability of \texttt{HeTriNet}. It demonstrates the model's adaptability and robustness across different datasets and reinforces its predictive capabilities in real-world scenarios. Our findings provide valuable insights into the model's practical utility and its potential to enhance our understanding of complex drug-target-disease interactions.
\vspace{-4mm}
\subsection{Case Study on Depression}
\vspace{-2mm}
In the field of medicine, one of our core objectives is to provide personalized treatments that cater to each patient's unique needs. To achieve this, it is crucial to identify which drugs are effective for specific diseases and understand the biological targets associated with disease pathways. In our research, we have harnessed the predictive capabilities of \texttt{HeTriNet} to uncover new combinations of drugs and biological targets relevant to depression – a complex condition with various molecular factors at play. Depression has a multifaceted origin. Our work with \texttt{HeTriNet} offers a promising way to revolutionize the treatment of depression and similar complex conditions.

\begin{table}[t]
    \centering
      \caption{Top 5 novel drug-target pairs for depression}
      \begin{tabular}{|c|c|c| }
      \hline
       \cellcolor{gray!60}Drug (DrugBank) & \cellcolor{gray!60}Target (UniProt) &
       \cellcolor{gray!60}Evidence
       \\
       \hline
Amitriptyline & Sodium-dependent serotonin transporter & Kim Lawson \cite{Lawson2017ABR}\\
Nortriptyline & 5-hydroxytryptamine receptor 2A & Pierre Blier\cite{Blier2013NeurotransmitterTI}\\
Imipramine & Sodium-dependent serotonin transporter & Dempsey et al.\cite{Dempsey2005SerotoninF}\\
Nortriptyline & Muscarinic acetylcholine receptor M5 & Philip et al.\cite{Philip2010NicotinicAR}\\ 
Nortriptyline & MD(2) dopamine receptor & Pierre Blier\cite{Blier2013NeurotransmitterTI}\\ 
        \hline
        \end{tabular} \vspace{-2mm}
        \label{Tab9}
\end{table}

Table~\ref{Tab9} enlists the top five novel pairs of (drug, target) corresponding to depression and literature evidence supporting these predictions. This result is remarkable because we were able to find literature evidence for these five drug-target pairs. In essence, \texttt{HeTriNet} represents an efficient way to find potential treatments for various diseases, starting with depression. It is a significant step toward personalized medicine, where treatments are tailored to each patient's unique needs. The model's ability to make reliable predictions supported by scientific evidence holds great promise for revolutionizing how we treat complex diseases.

\vspace{-4mm}
\subsection{Ablation Study}
\vspace{-2mm}
To validate the effectiveness of each component in \texttt{HeTriNet}, we design four different variants for the ablation study:

\begin{itemize}
\item \texttt{HeTriNet-Sum}: Utilizes summation on neighbor node embeddings in Eq~\ref{eq4} instead of concatenation and neural network transformations.

\item \texttt{HeTriNet-Concat}: Concatenates three neighbor node embeddings and then reduces their dimension without neural network transformations.

\item \texttt{HeTriNet-Elem-Prod}: Applies element-wise product on neighbor node embeddings in Eq ~\ref{eq4}.

\item \texttt{HeTriNet-Trans}: Concatenates three neighbor node embeddings and then reduces their dimension without neural network transformations.

\end{itemize}

\begin{wrapfigure}{r}{0.55\textwidth}\vspace{-8mm}
    \centering
    \includegraphics[width=0.49\textwidth]
    {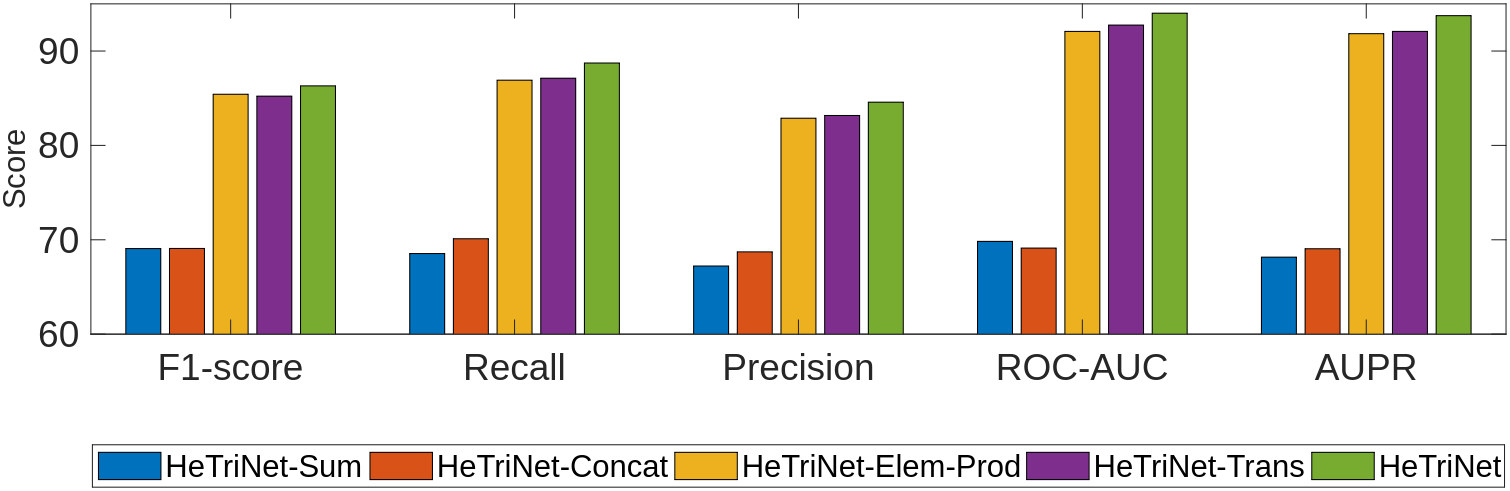}
   \caption{Performance Comparison of \texttt{HeTriNet} with its variants}
    \label{fig:HeTriNet variants} 
\vspace{-5mm}
\end{wrapfigure}

Comparing these variants with the original \texttt{HeTriNet}, we observe that \texttt{HeTriNet-Sum} and \texttt{HeTriNet-Concat} have poorer performance than \texttt{HeTriNet}, likely due to their use of summation or concatenation for neighbor node embedding. 
\texttt{HeTriNet-Elem-Prod} and \texttt{HeTriNet-Trans} also exhibit lower performance than \texttt{HeTriNet}, emphasizing the effectiveness of applying a neural network to concatenated node embeddings. Table~\ref{Tab9} outlines detailed performance comparisons on triplet prediction tasks, which further supports the efficacy of the proposed triplet-attention mechanism in a heterogeneous graph.

\vspace{-4mm}
\section{Conclusion}
\label{conclusion}
\vspace{-2mm}
In conclusion, our study introduces the Heterogeneous Graph Triplet Attention Network (\texttt{HeTriNet}) as a powerful model for modeling drug-target-disease interactions. We have addressed the limitations of existing methods by proposing a dedicated heterogeneous graph neural network that leverages the triplet attention mechanism to improve prediction accuracy. The concepts and techniques introduced in our work, including the triplet attention mechanism, have broad generalizability and applicability. They can be extended to various domains and problems involving heterogeneous graphs and higher-order interactions.

Looking ahead, future research directions could explore the incorporation of additional factors such as drug-target-pathway-disease interactions, enabling a deeper understanding of drugs' mechanism of action (MoA) and facilitating drug discovery. Moreover, integrating multi-omics data and exploring more complex heterogeneous graph structures hold promise for enhancing the predictive power of our model and providing a more comprehensive view of underlying biological processes. These advancements have the potential to drive progress in personalized medicine and targeted drug development, ultimately benefiting patients and healthcare systems alike.

\vspace{-4mm}
\bibliography{sample}

\end{document}